\documentclass[journal]{IEEEtran}
\usepackage{amsmath,amssymb}
\usepackage{graphicx}
\usepackage[ruled]{algorithm2e}
\usepackage{cite}

\newcommand{\tabincell}[2]{\begin{tabular}{@{}#1@{}}#2\end{tabular}}

\begin{document}

\title{A Neural Network Aided Approach for LDPC Coded DCO-OFDM with Clipping Distortion}

\author{
Yuan~He,~\IEEEmembership{Student Member,~IEEE,} Ming~Jiang,~\IEEEmembership{Member,~IEEE,}
and~Chunming~Zhao,~\IEEEmembership{Member,~IEEE}

\thanks{Yuan He, Ming Jiang, *** and Chunming Zhao are with the National Mobile Communications Research Laboratory, Southeast University, Nanjing 210096, China (e-mail:\{heyuan, jiang\_ming, xtling, cmzhao\}@seu.edu.cn).}

\thanks{This work was supported by the National Natural Science Foundation of China (61771133, 61521061 and 61601115), the National Science \& Technology Projects of China under grant 2018ZX03001002.}
}

\maketitle

\begin{abstract}

In this paper, a neural network-aided bit-interleaved coded modulation (NN-BICM) receiver is designed to mitigate the nonlinear clipping distortion in the LDPC coded direct current-biased optical orthogonal frequency division multiplexing (DCO-OFDM) systems. Taking the cross-entropy as loss function, a feed forward network is trained by backpropagation algorithm to output the condition probability through the $softmax$ activation function, thereby assisting in a modified log-likelihood ratio (LLR) improvement. To reduce the complexity, this feed-forward network simplifies the input layer with a single symbol and the corresponding Gaussian variance instead of focusing on the inter-carrier interference between multiple symbols. On the basis of the neural network-aided BICM with Gray labelling, we propose a novel stacked network architecture of the bit-interleaved coded modulation with iterative decoding (NN-BICM-ID). Its performance has been improved further by calculating the condition probability with the aid of \(a\) \(priori\) probability that derived from the extrinsic LLRs in the LDPC decoder at the last iteration, at the expense of customizing neural network detectors at each iteration time separately. Utilizing the optimal DC bias as the midpoint of the dynamic region, the simulation results demonstrate that both the NN-BICM and NN-BICM-ID schemes achieve noticeable performance gains than other counterparts, in which the NN-BICM-ID clearly outperforms NN-BICM with various modulation and coding schemes.


\end{abstract}

\section{Introduction}

Visible light communications (VLC) have become an emerging short-range communication technique for the indoor scenarios to complement the radio frequency (RF) systems \cite{Rajagopal2012}. With such distinct advantages as the abundant unlicensed spectrum, low cost and security, VLC systems can support the communication and illumination simultaneously by adopting the intensity modulation and direct detection (IM/DD) to guarantee the real non-negativity for driving the light emitting diode (LED). To achieve a higher transmission rate, the optical orthogonal frequency division multiplexing (OFDM) has attracted much attention for the multi-carrier VLC applications in comparison with the single-carrier pulse modulation schemes, e.g. the on-off keying (OOK) and pulse position modulation (PPM), due to its spectral efficiency and robustness against the inter symbol interference (ISI).

In multi-carrier VLC systems, there are many variants of optical OFDM modulation schemes to generate the real and non-negative intensity signals \cite{Armstrong2013,Wang2014}. Particularly, DCO-OFDM exhibits the highest spectral efficiency with simple implementations, in which the Hermitian symmetry can ensure the real-valued property and the DC bias can handle the non-negativity constraint \cite{Ling2016}. For the optical front-end, the transfer characteristic of an LED after the pre-distortion can be modeled as a dynamic-range-limited nonlinearity, where the linear dynamic range is limited between the minimum and maximum input current \cite{Haas2013,Ying2015WC}. The double-sided clipping should be adopted to accommodate DCO-OFDM signals within the dynamic range constraint. However, the DCO-OFDM signals with high peak-to-average power ratio (PAPR) show a considerable sensitivity to the nonlinear distortion caused by the double-sided clipping operation inevitably. The efficient methods to mitigate the nonlinear distortion are the bit-interleaved coded modulation (BICM) receivers combined with the clipping nonlinearity. Taking advantage of the near Shannon performance and high throughput iterative decoding, BICM potentially chooses the low density parity check (LDPC) coding scheme to exhibit a significant robustness to the impulsive interference \cite{Zhang2015,Ivan2012}.

In the context of the conventional LDPC coded BICM receivers, the maximum \(a\) \(posteriori\) (MAP) demapper derives the mismatched extrinsic log likelihood ratio (LLR) values due to the nonlinear inter-carrier distortion caused by double-sided clipping operations, resulting in a serious degradation. The MAP-BICM is the BICM receiver based on the MAP detection with the assumption of the Gaussian noise, which suffers from the mismatched soft output when the clipping distortion incurs. Most previous works have focused on the improved BICM designs based on the clipping distortion, mainly consisting of BICM receiver based on maximum sequence likelihood (MSL-BICM), BICM receiver based on Gaussian mixture model (GMM-BICM) and so on \cite{Tan2016,Lyu2015}. For example, MSL-BICM is an enhanced near-optimal BICM design for the clipped DCO-OFDM system by revising the LLR criterion based on the maximum sequence likelihood \cite{Tan2016}. Since the revised LLR criterion consumes extra complexity with increasing subcarriers and suffers from the imperfect channel state information (CSI), MSL-BICM shall be limited by inter-carrier distortion between the numerous subcarriers. GMM-BICM models the channel conditional probability that the equalizer outputs as the mix-Gauss distribution and obtains the modified LLR values \cite{Lyu2015}. Despite several advantages including modeling the probability distributions with any required accuracy level and convenience of using the expectation maximization (EM) algorithm, GMM are statistically inefficient for modeling in a nonlinear manifold of the data space \cite{Hinton2012}. It leads to a limitation in performance when using the GMM to model channel conditional probability after the clipping operation.

Recently, machine learning (ML) has attracted growing interest in the potential applications of the physical layer, including channel estimation and detection, equalization  and channel decoding etc. \cite{Jin2017,zhang2018,Samuel2017,Li2017,Nachmani2018}. In \cite{Samuel2017}, the authors propose a deep learning-based maximum likelihood detector, named DetNet, with a unfolding architecture by adopting the projected gradient descent algorithm. Besides the robustness to the imperfect CSI, a neural network (NN) detector in \cite{Li2017} is expected to learn a much better model with the data in a nonlinear manifold. Firstly, the concept of symbol-by-symbol detection and sequence detection are put forward \cite{Farsad2018}, and the authors established a framework of NN with the cross-entropy loss function following the $softmax$ activation function to output the probability of the estimations. Several works have discussed the similar methods \cite{Dorner2018,Oshea2017,Oshea2017-2}. These advances trigger interest in developing the BICM receiver based on the NN, where the layered neuron model aims at recovering the desired transmitted symbols from the corrupted signals.

Motivated by this goal, we propose a reduced complexity NN-aided BICM receiver for the LDPC coded DCO-OFDM system, in which a feed-forward NN is trained to learn the channel condition probability. Instead of focusing on the inter-carrier distortion between multiple received symbols, this basic NN architecture simplifies the input of the single symbol and corresponding AWGN variance to reduce complexity. With the cross-entropy loss function, the NN is trained by backpropagation algorithm to output the condition probability through the $softmax$ activation function, thereby assisting in the LLR improvement. The rest of this paper is organized as follows. In Section II, we propose a reduced complexity NN-aided BICM receiver design for the LDPC coded DCO-OFDM system. Specifically, we present a hybrid architecture and implementation for the receiver with NN in Section III, including computational complexity and improvement of overfitting. Utilizing the optimal DC bias, simulation results demonstrate that the NN-aided BICM receiver with different modulation and coding schemes, compared with other counterparts in Section IV. Conclusions are drawn in Section V.

\begin{figure}[!t]
		\setlength{\abovecaptionskip}{0pt}
		\setlength{\belowcaptionskip}{0pt}
		\centering
		\includegraphics[ width=90mm]{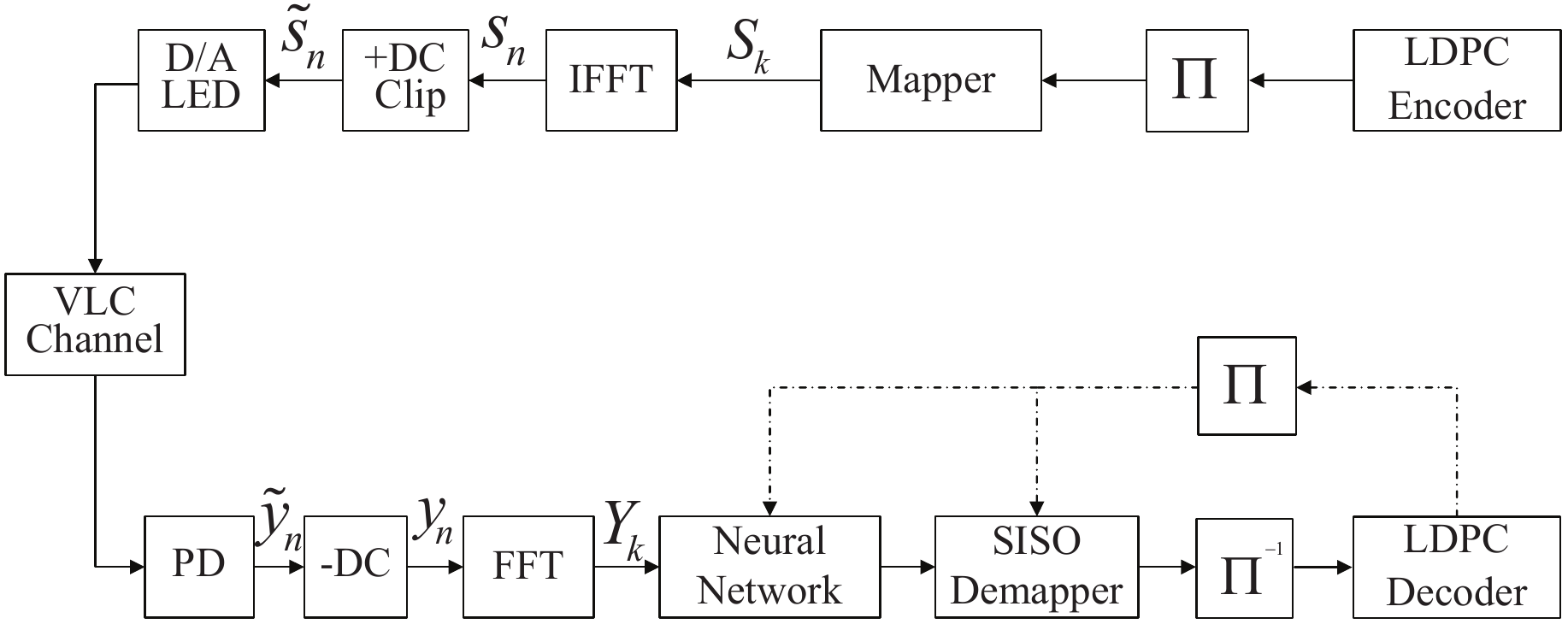}
		\caption{Block diagram of the LDPC coded DCO-OFDM system employing a NN-aided BICM receiver.}
		\label{fig1} \end{figure}

\section{System Model}

Fig.1 shows an LDPC coded DCO-OFDM system combined with the NN-aided BICM receiver. At the transmitter, an rate-\(R\) LDPC encoder encodes the independent bit streams. To break the fading correlation, the coded bit streams are permuted by a quasi-random interleaver \(\Pi\). Consider the labelling rules, each \(M\) interleaved bit streams are mapped onto a modulated \({2^M}\)-QAM symbol in the \({2^M}\)-ary constellation set \(\chi \). BICM can be defined as a concatenation of the rate-\(R\) LDPC encoder with \({2^M}\)-ary memoryless modulator, which is separated by the interleaver \(\Pi\).

For driving the LED, the transmitter enjoys the intensity electrical signals with real and non-negative properties via an intensity modulation and direct detection (IM/DD) scheme. Specifically, the information-carrying symbols ${\left[{S_1},...,{S_{{N/2}-1}} \right]}$ are allocated over \(N\) subcarriers by following the Hermitian symmetry ${S_k} = S_{N - k}^*, k = 1,..., N/2-1$, except that the $0$-th and $N/2$-th ones being set to zero. The real-valued time-domain signals ${\left[{s_0},...,{s_{N-1}} \right]}$ can be obtained by an \(N\) point inverse fast Fourier transform (N-IFFT) at the expense of \(50\% \) reduction in spectral efficiency, as follows
\begin{align}
{s_n} = \frac{1}{{\sqrt N }}\sum\limits_{k = 0}^{N - 1} {{S_k}{e^{j\frac{{2\pi nk}}{N}}}},\ n = 0, ..., N - 1.\label{equ1}
\end{align}

Due to a dynamic-range constraint on the LED, the DCO-OFDM signals ${s_n}$ are biased with a DC bias ${\mu}$ and the resulting double side clipping regarding the clipped signals ${\tilde s_n}$ can be expressed as
\begin{align}
{\tilde s_n} = \left\{ \begin{array}{l}
{\Omega _b},{\kern 1pt} {\kern 1pt} {\kern 1pt} {\kern 1pt} {\kern 1pt} {\kern 1pt} {\kern 1pt} {\kern 1pt} {\kern 1pt} {\kern 1pt} {\kern 1pt} {\kern 1pt} {\kern 1pt} {\kern 1pt} {\kern 1pt} {\kern 1pt} {\kern 1pt} {\kern 1pt} {\kern 1pt} {\kern 1pt} {\kern 1pt} {\kern 1pt} {\kern 1pt} {\kern 1pt} {\kern 1pt} {\kern 1pt} {\kern 1pt} {\kern 1pt} {\kern 1pt} {\kern 1pt} {\kern 1pt} {\kern 1pt} {\kern 1pt} {\kern 1pt} {\kern 1pt} {\kern 1pt} {\kern 1pt} {\kern 1pt} {\kern 1pt} {\kern 1pt} {\kern 1pt} {\kern 1pt} {\kern 1pt} {\kern 1pt} {\rm if} {s_n}  \le {\Omega _b}-\mu\\
{s_n} + \mu ,{\kern 1pt} {\kern 1pt} {\kern 1pt} {\kern 1pt} {\kern 1pt} {\kern 1pt} {\kern 1pt} {\kern 1pt} {\kern 1pt} {\kern 1pt} {\kern 1pt} {\kern 1pt} {\kern 1pt} {\kern 1pt} {\kern 1pt} {\kern 1pt} {\kern 1pt} {\kern 1pt} {\kern 1pt} {\kern 1pt} {\kern 1pt}  {\kern 1pt} {\kern 1pt} {\kern 1pt} {\kern 1pt} {\kern 1pt} {\kern 1pt} {\rm if} {\Omega _b}-\mu < {s_n}  \le {\Omega _t}-\mu \\
{\Omega _t},{\kern 1pt} {\kern 1pt} {\kern 1pt} {\kern 1pt} {\kern 1pt} {\kern 1pt} {\kern 1pt} {\kern 1pt} {\kern 1pt} {\kern 1pt} {\kern 1pt} {\kern 1pt} {\kern 1pt} {\kern 1pt} {\kern 1pt} {\kern 1pt} {\kern 1pt} {\kern 1pt} {\kern 1pt} {\kern 1pt} {\kern 1pt} {\kern 1pt} {\kern 1pt} {\kern 1pt} {\kern 1pt} {\kern 1pt} {\kern 1pt} {\kern 1pt} {\kern 1pt} {\kern 1pt} {\kern 1pt} {\kern 1pt} {\kern 1pt} {\kern 1pt} {\kern 1pt} {\kern 1pt} {\kern 1pt} {\kern 1pt} {\kern 1pt} {\kern 1pt} {\kern 1pt} {\kern 1pt} {\kern 1pt} {\kern 1pt} {\kern 1pt} {\rm if} {s_n}   > {\Omega _t}-\mu
\end{array} \right.
\end{align}
where the top and bottom clipping levels denote ${\Omega _t}$ and ${\Omega _b}$ respectively. According to the Bussgang theorem, the clipped signals can also be calculated by
\begin{align}
{\tilde s_n} = \alpha {s_n} + {d_n} ,{\kern 1pt} {\kern 1pt} {\kern 1pt} {\kern 1pt} {\kern 1pt} {\kern 1pt} n = 0,...,N - 1
\end{align}
where \(\alpha \) is the attenuation factor and ${d_n}$ is the clipping distortion \cite{Bussgang1952}. The attenuation factor equals to
\begin{align}
\alpha  = Q({\phi _b} - \phi ) - Q({\phi _t} - \phi )
\end{align}
where $Q\left( \phi \right)$ represents the Gaussian Q-function of $\phi$ \cite{Haas2013}. In addition, $\phi$ is the ratio of DC bias and signal power $\phi = \mu/\sigma_s$ over a range of the minimum value ${\phi _b}$ and maximum ones ${\phi _t}$, i.e., ${\phi _b} = {\Omega _b}/{\sigma _s}$ and ${\phi _t} = {\Omega _t}/{\sigma _s}$. The clipped signal $\tilde s_n$ drives the intensity of a LED to generate the visible light signal \({s_{\mu}}(t)\). In \cite{Armstrong2013}, the electrical power of the transmitted signals \({s_{\mu}}(t)\) can be evaluated by
\begin{align}
\begin{array}{l}
{P_e}(\delta _b,\delta _t ,{\sigma _s}) = \sigma _s^2(Q({\delta _b}) - Q({\delta _t}) + {\delta _b}g({\delta _b}) - {\delta _t}g({\delta _t})\\
{\kern 1pt} {\kern 1pt} {\kern 1pt} {\kern 1pt} {\kern 1pt} {\kern 1pt} {\kern 1pt} {\kern 1pt} {\kern 1pt} {\kern 1pt} {\kern 1pt} {\kern 1pt} {\kern 1pt} {\kern 1pt} {\kern 1pt} {\kern 1pt} {\kern 1pt} {\kern 1pt} {\kern 1pt} {\kern 1pt} {\kern 1pt} {\kern 1pt} {\kern 1pt} {\kern 1pt} {\kern 1pt} {\kern 1pt} {\kern 1pt} {\kern 1pt} {\kern 1pt} {\kern 1pt} {\kern 1pt} {\kern 1pt} {\kern 1pt} {\kern 1pt} {\kern 1pt} {\kern 1pt} {\kern 1pt} {\kern 1pt} {\kern 1pt} {\kern 1pt} {\kern 1pt} {\kern 1pt} {\kern 1pt} {\kern 1pt} {\kern 1pt}  + \delta _b^2Q( - {\delta _b}) + \delta _t^2Q({\delta _t}))
\end{array}.\label{equ5}
\end{align}
where the standardized normal distribution $g(\phi )$ equals to $g(\phi ) = \frac{1}{{\sqrt {2\pi } }}\exp ( - \frac{{{\phi ^2}}}{2})$, the difference values ${\delta _b}$ and ${\delta _t}$ denote ${\phi _b} - \phi $ and ${\phi _t} - \phi$ respectively.

Typically, the VLC channel can be modeled as a low-pass time-invariant channel plus the AWGN noise. The corresponding received optical signal is converted into the  electrical signal $\tilde y_n$ by the the photodiode (PD). After performing \(N\) point fast Fourier transform (N-FFT), the signals $y_n$ are transformed into the frequency domain symbols $Y_k$. The symbols $Y_k$ contains the attenuated symbol $S_k$ with factor \(\alpha \) and the channel frequency response ${H_k}$, the clipping distortion $D_{k}$ and the AWGN noise ${W_k}$ with zero mean and variance \({\sigma_{n}^2}\) on the \(k\)-th OFDM subcarrier respectively \cite{Zhang2015}, i.e.,
\begin{align}
{Y_k} = \alpha {H_k}{S_k} + {H_k}{D_{k}} + {W_k}.
\end{align}
For convenience, the channel response is normalized as \({H_k} = 1,\forall k\).

The NN-aided BICM receiver is trained to learn the condition probability $p({Y_k}|S_k)$, where the received signal $Y_k$ suffers from a clipping distortion. This network adopts the fully connected feed-forward architecture followed by an output layer with the $softmax$ activation function, in which the \(L-1\) hidden layers with $\rm tanh$ activation function can be chosen. Here, the $\rm tanh$ function is written as
\begin{align}
{\rm tanh}(x_j^{(l)}) = \frac{2}{{1 + {e^{ - 2x_j^{(l)}}}}} - 1
\end{align}
$d^{(l)}$ and $x_j^{(l)}$ denote the number of neurons and the $j$-th neuron in the layer $l$ respectively, $1 \le l \le L-1$, $1 \le j \le d^{(l)}$. The input layer contains ${d^{(0)}} = 3$ elements, including the real and imaginary components ${\rm Re} \{ {Y_k}\}$, ${\rm Im} \{ {Y_k}\}$ and the variance \({\sigma_{n}^2}\). The $L$-th layer outputs the condition probability $p({Y_k}|S_k)$ through the $softmax$ function, in which ${d^{(L)}}$ equals to size of the \({2^M}\)-ary constellation set. The NN is carried out with the cross-entropy loss function. For the gradient descent optimization, the backpropagation algorithm aims at training such a network efficiently with the scaled conjugate gradient (SCG) method. The NN can assist in the LLR calculation by obtaining the sum of the probability ${p({Y_k}|S_k^m)}$ in set of the symbol ${S_k}$ whose the \(m\)-th bit equals to $b$.

In the maximum \(a\) \(posteriori\) (MAP) demapper, the LLR $L_{\rm DEM}^{k,m}$ of the m-th bit in symbol $S_k, k = 1,..., N/2-1$ is calculated by equation \cite{Tan2016}
\begin{align}
L_{\rm DEM}^{k,m} = \log \frac{{\sum\limits_{S_k^m \in {\bf{\chi }}_1^m} {p({Y_k}|S_k^m)} }}{{\sum\limits_{S_k^m \in {\bf{\chi }}_0^m} {p({Y_k}|S_k^m)} }},{\kern 1pt} {\kern 1pt} {\kern 1pt} m = 1,...,M, \label{equ7}
\end{align}
where ${\chi }_b^{m}$ stands for the set of symbols whose the \(m\)-th bit is $b=0, 1$. The LLRs are passed through a quasi-random deinterleaver \(\Pi^{-1}\) and sent to the LDPC decoder. The iterative decoding of LDPC codes can be viewed as a serial concatenation with an inner variable-node decoder (VND) and an outer check-node decoder (CND). The extrinsic LLRs between the VND and CND are iteratively updated to form the final decisions until all the parity-check equations are satisfied or the maximum number of iterations is reached \cite{StBrink2004}.

\section{Hybrid BICM Receiver with NN }

\subsection{Network Architecture and Implementation}
In the conventional BICM receivers, the conditional probability $p({Y_k}|S_k;\theta)$ with clipping parameters set $\theta  = \{ \alpha ,{H_k},{D_k}\}$ is usually formulated as Gaussian model, which does not accurately characterizes the effect of the clipping distortion. Moreover, for the clipped DCO-OFDM, the inter-carrier distortion caused by nonlinearity distortion in MAP demodulator makes the error distribution function extremely complicated than the Gaussian assumption. Several research works devote to improving the demodulation performance with the design of NN \cite{zhang2018,Samuel2017,Li2017}, but it still remains a challenge on the soft decision calculation after the NN.

\begin{figure}[!t]
		\setlength{\abovecaptionskip}{0pt}
		\setlength{\belowcaptionskip}{0pt}
		\centering
		\includegraphics[ width=90mm]{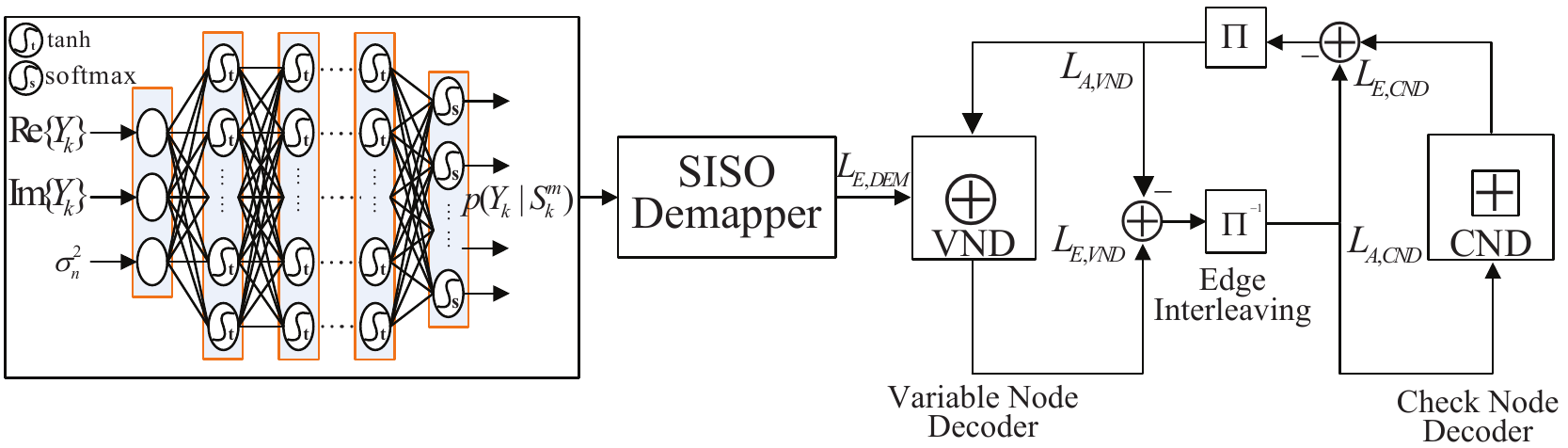}
		\caption{Hybrid LDPC coded BICM receiver with the NN.}
		\label{fig2} \end{figure}

In \cite{Lyu2015}, the authors exploit the discriminative strategy, such as GMM, to model the error distribution and get the soft output. However, the potential of GMM techniques is restricted by the limitations of the discriminative model, in which the estimated GMM parameters for each \(E_b/N_0\) require extensive iterative computations. We believe that the feedforward NN with a simplified structure offers a dramatic capability to model the nonlinear function $p({Y_k}|S_k;\theta)$ by training the network weights with error derivatives from back propagation algorithm. Theoretically, the feedforward NN with three layers has been proven to approximate any continuous nonlinear function with an arbitrary degree of precision, in the condition of enough hidden neurons \cite{Hornik1989}. In \cite{Oshea2017}, the feedforward NN belongs to an undercomplete denoising autoencoder that is trained to reconstruct the original data from the corrupted inputs, as long as the encoder function is deterministic. In this paper, we propose a hybrid design of the NN-aided BICM receiver, in which the feedforward NN learns the embedded conditional probability and the soft input soft output (SISO) demapper derives its LLR value. This hybrid design is reasonable for combining the generalization ability on the NN and the inferential capability on the Bayesian network.

Different from the multi-symbol design of the input layer in other structures, the proposed NN only takes a single received symbol $Y_k, k = 1,..., N/2-1$ and the corresponding variance \({\sigma_{n}^2}\) as input, instead of the inter-carrier interference (ICI) cancellation among multiple subcarriers. For the DCO-OFDM, ICI can be expressed as a polynomial nonlinear function of \(N\) complex symbols $S_k$, in which the NN for ICI elimination would consume numerous neurons with extra layers []. Instead, our network only aims at establishing a nonlinear analytic function between $Y_k$ and $S_k$ by considering the noise variance and clipping distortion. Notice that whether the input includes variance \({\sigma_{n}^2}\) or not will have a strong influence on the performance, because the Gaussian noise can help the network optimization. As a consequence, our network can decrease the size of network architecture by employing fewer layers, thereby simplifying the optimization process and reducing the computation complexity.

The implementation of the NN-aided receiver contains the training and testing procedure. In the training procedure, we train the NN using a data-driven strategy when given the modulation order \(M\), code rate \(R\) and electrical training \(E_b/N_0\) \(\gamma_{e}\). The training dataset $\varphi$ takes the recieved symbol $Y_k$ and the corresponding variance \({\sigma_{n}^2}\) as input, and the probability mass function (PMF) $p_{\varphi} ({S_k})$ of the transmitted symbol $S_k$ that defined on the \({2^M}\)-ary space \(\chi \) as target respectively. PMF $p_{\varphi} ({S_k})$ takes the form of indicator function, i.e.,
\begin{align}
{p_{\varphi}}({S_k}) = [{\rm{{\cal I}}}\{ {S_k} = {\chi_1}\} {\kern 1pt} {\kern 1pt} {\rm{{\cal I}}}\{ {S_k} = {\chi_2}\}  \cdots {\rm{{\cal I}}}\{ {S_k} = {\chi_{2^M}}\} ]
\end{align}
where ${\rm{{\cal I}}}\{\cdot \}$ denotes the indicator function, ${\chi_j}$ is a specific constellation point in the \(j\)-th labelling index and the maximum value of the labelling index satisfies  $\max (j) = 2^M$.

To output the soft decision, the NN converts the linear aggregation \({x_{j}^{(L)}}\) of inputs \({a_{i}^{(L-1)}}\) from the previous \((L-1)\)-th layers into the posterior probability ${p_\varphi }({S_k}|Y_k;\theta)$ by using the $softmax$ activation function \cite{Farsad2018},
\begin{align}
{p_\varphi }({S_j}|{Y_k};\theta ) = \frac{{\exp ({x_{j}^{(L)}})}}{{\sum\limits_j {\exp ({x_{j}^{(L)}})} }}
\end{align}
It's also noted that ${p_\varphi }(S_k)$ is the prior probability of $S_k$ in the training set $\varphi$, ${p_\varphi }({S_k}|Y_k;\theta)$ is the posterior probability that the NN outputs.

Like the multiclass classification, the output layer uses the $softmax$ function to derive the posterior probability ${p_\varphi }({S_k}|Y_k;\theta)$ and the loss function $J(\omega;\theta )$ can be chosen as cross-entropy between the target PMF $p_{\varphi} ({S_k})$ and the output of the $softmax$ ${p_\varphi }({S_k}|Y_k;\theta)$, given by
\begin{align}
J(\omega;\theta ) =  - \frac{1}{{\left| \varphi  \right|}}\sum\limits_{k \in \varphi } {\sum\limits_{j \in \chi } {{\rm{{\cal I}}}\{ {S_k} = {S_j}\} } } \log {p_\varphi }({S_j}|{Y_k};\theta).
\label{equ11}
\end{align}
Thus, the NN weights are fine-tuned by optimizing the cross-entropy, which is equivalent to the maximum likelihood principle. According to the Bayes' theorem, the conditional probability ${p_\varphi }({Y_k}|S_k;\theta)$ equals to ${p_\varphi }({S_k}|Y_k;\theta)$, suppose ${p_\varphi }(S_k)$ is uniformly distributed. Then, the NN can produce the conditional probability $p({Y_k}|S_k;\theta)$, because the output ${p_\varphi }({S_k}|Y_k;\theta)$ can be converted into the likelihood ${p_\varphi }({Y_k}|S_k;\theta)$. For the testing procedure, the MAP demapper can exploit the probability ${p_\varphi }({Y_k}|S_k;\theta)$ that the NN outputs to calculate the LLR $L_{\rm DEM}^{k,m}$ by the equation (\ref{equ7}), where the following steps have been discussed in section II.

\begin{algorithm}[t!]
\small
\caption{Hybrid BICM receiver with NN}
{
{\bf Training procedure:} \\
\KwIn{symbols \({Y_k} \in \varphi \), \(E_b/N_0\) \(\gamma_t\), PMF $p_{\varphi} ({S_k})$}
\KwOut{weights {\bf \(\omega\)}}
\For{all $({Y_k},\sigma_n^2,{p_\varphi }({S_k})) \in \varphi $}{
　　Initialize weights $w_{ij}^{(l)}$ and biases $w_{0j}^{(l)}$ $\in (0,1)$ randomly\;
    Get $\sigma_n^2$ according to \(\gamma_t\)\;
　　\For{$1 \le l \le L$}{
　　Calculate outputs $a_{i}^{(l)}$ from Eq.(\ref{equ12}) and Eq.(\ref{equ13})\;}
    Calculate cross-entropy $J(\omega;\theta )$ from Eq.(\ref{equ11})\;
    Derive gradient derivatives from Eq.(\ref{equ19}) and Eq.(\ref{equ20})\;
    Update weights by Eq.(\ref{equ22}) with learning rate \(\eta\)\;
}
~\\
{\bf Testing procedure:} \\
\KwIn{${\rm Re} \{ {Y_k}\}$, ${\rm Im} \{ {Y_k}\}$, $\sigma_n^2$}
\KwOut{$p({Y_k}|S_k;\theta)$}
\For{$1 \le l \le L$}{
　　Calculate outputs $p({Y_k}|S_k;\theta)$ from Eq.(\ref{equ12}) and Eq.(\ref{equ13})\;}
    Calculate LLRs $L_{\rm DEM}^{k,m}$ from Eq.(\ref{equ7})
}
\end{algorithm}

\begin{algorithm}[t!]
\small
\caption{Stacked BICM-ID receiver with NN}
{
{\bf Training procedure:} \\
\While{iteration $\le$ max}{
　　\eIf{iteration=1}{
        \KwIn{symbols \({Y_k} \in \varphi \), \(E_b/N_0\) \(\gamma_t\), PMF $p_{\varphi} ({S_k})$}
        \KwOut{weights {\bf \(\omega\)}}
　　　　\For{all $({Y_k},\sigma_n^2,{p_\varphi }({S_k})) \in \varphi $}{
　　Train net1 the same as Algorithm 1\;}}
   {
    \KwIn{symbols \({Y_k} \in \varphi \), \(E_b/N_0\) \(\gamma_t\), PMF $p_{\varphi} ({S_k})$, $p(S_k)$}
    \KwOut{weights {\bf \(\omega\)}}
　　\For{all $({Y_k},\sigma_n^2,p(S_k),{p_\varphi }({S_k})) \in \varphi $}{
　　Train net2 the same as Algorithm 1\;}
　　}
iteration=iteration+1\;
}

~\\
{\bf Testing procedure:} \\

\While{iteration $\le$ max}{
　　\eIf{iteration=1}{\KwIn{${\rm Re} \{ {Y_k}\}$, ${\rm Im} \{ {Y_k}\}$, $\sigma_n^2$}
    \KwOut{$p({Y_k}|S_k;\theta)$}
    \For{$1 \le l \le L_1$}{
　　Calculate net1 outputs $p({Y_k}|S_k;\theta)$ from Eq.(\ref{equ12}) and Eq.(\ref{equ13})\;}
    Calculate LLRs $L_{\rm DEM}^{k,m}$ from Eq.(\ref{equ7})\;
    Calculate $p(S_k)$ from feed back $L_{A,DEM}^{k,m}$ by Eq.(\ref{equ24}) and return $p(S_k)$ back to the training procedure\;
  }{
  \KwIn{${\rm Re} \{ {Y_k}\}$, ${\rm Im} \{ {Y_k}\}$, $\sigma_n^2$, $p(S_k)$}
  \KwOut{$p({Y_k}|S_k;\theta)$}
  \For{$1 \le l \le L_2$}{
　　Calculate net2 outputs $p({Y_k}|S_k;\theta)$ from Eq.(\ref{equ12}) and Eq.(\ref{equ13})\;}
    Calculate LLRs $L_{\rm DEM}^{k,m}$ from Eq.(\ref{equ23})\;}
    iteration=iteration+1\;
  }

}
\end{algorithm}

\subsection{Backpropagation Performance Analysis}

Here, we consider the numerical analysis of feedforward network for minimizing the loss function $J(\omega;\theta )$ with weights trained by backpropagating error derivatives. The inputs are propagated through the neuron layer by layer in the forward pass to generate the outputs. By calculating the gradient of the loss function, the resulting error derivatives are fed back to adjust weights iteratively in the hidden layers by the gradient descent. For each layer, the neuron takes the nonlinear activation function of the weighted combination $x_j^{(l)}$ with respect to its inputs $a_i^{(l - 1)}$,

\begin{itemize}
  \item \textbf{Linear aggregation}

\begin{align}
x_j^{(l)} = \sum\limits_{i = 0}^{{d^{(l - 1)}}} {w_{ij}^{(l)}a_i^{(l - 1)}},{\kern 1pt} {\kern 1pt} {\kern 1pt} 1 \le l \le L
\label{equ12}
\end{align}

\end{itemize}

\begin{itemize}
  \item \textbf{Activation function}

\begin{align}
a_j^{(l)} = \left\{ \begin{array}{l}
\tanh (x_j^{(l)}),{\kern 1pt} {\kern 1pt} {\kern 1pt} 1 \le l \le L - 1\\
softmax(x_j^{(l)}),{\kern 1pt} {\kern 1pt} {\kern 1pt} l = L
\end{array} \right.
\label{equ13}
\end{align}

\end{itemize}
where $w_{ij}^{(l)}$ means the weight value from the \(i\)-th input in \((l-1)\)-th layer to the \(j\)-th input in \(l\)-th layer and $d^{(l)}$ denotes the number of the neurons. Notice that the \(\rm tanh\) and \(softmax\) function are used in the hidden and output layers, respectively.

Backpropagation is an automatic differentiation technique used to adjust the weights by following the gradient-based optimization with error derivatives of $J(\omega;\theta )$ \cite{Bishop2006}. In the output layer, the gradient derivative can be obtained by the chain rule,

\begin{align}
\frac{{\partial J(\omega ;\theta )}}{{\partial w_{ij}^{(L)}}} = \sum\limits_{i = 0}^{{d^{(L)}}} {\frac{{\partial J(\omega ;\theta )}}{{\partial a_i^{(L)}}}\frac{{\partial a_i^{(L)}}}{{\partial x_j^{(L)}}}} \frac{{\partial x_j^{(L)}}}{{\partial w_{ij}^{(L)}}}.
\end{align}
We can compute each factor in multiplication as

\begin{align}
\begin{array}{l}
\frac{{\partial J(\omega ;\theta )}}{{\partial a_i^{(L)}}} = \frac{\partial }{{\partial a_i^{(L)}}}( - \sum\limits_{j \in {d^{(L)}}} {{\rm{{\cal I}}}\{ {S_k} = {S_j}\} loga_j^{(L)}} )\\
{\kern 1pt} {\kern 1pt} {\kern 1pt} {\kern 1pt} {\kern 1pt} {\kern 1pt} {\kern 1pt} {\kern 1pt} {\kern 1pt} {\kern 1pt} {\kern 1pt} {\kern 1pt} {\kern 1pt} {\kern 1pt} {\kern 1pt} {\kern 1pt} {\kern 1pt} {\kern 1pt} {\kern 1pt} {\kern 1pt} {\kern 1pt} {\kern 1pt} {\kern 1pt} {\kern 1pt} {\kern 1pt} {\kern 1pt} {\kern 1pt} {\kern 1pt} {\kern 1pt} {\kern 1pt} {\kern 1pt}  =  - {\rm{{\cal I}}}\{ {S_k} = {S_i}\} \frac{1}{{a_i^{(L)}}}
\end{array}
\end{align}

\begin{align}
\begin{array}{l}
\frac{{\partial a_i^{(L)}}}{{\partial x_j^{(L)}}} = \frac{\partial }{{\partial x_j^{(L)}}}\frac{{\exp (x_i^{(L)})}}{{\sum\limits_{i = 0}^{{d^{(L)}}} {\exp (x_i^{(L)})} }}\\
{\kern 1pt} {\kern 1pt} {\kern 1pt} {\kern 1pt} {\kern 1pt} {\kern 1pt} {\kern 1pt} {\kern 1pt} {\kern 1pt} {\kern 1pt} {\kern 1pt} {\kern 1pt} {\kern 1pt} {\kern 1pt} {\kern 1pt} {\kern 1pt} {\kern 1pt} {\kern 1pt} {\kern 1pt} {\kern 1pt} {\kern 1pt} {\kern 1pt} {\kern 1pt}  = a_j^{(L)}(1 - a_j^{(L)}){\rm{{\cal I}}}\left\{ {j = i} \right\} - a_i^{(L)}a_j^{(L)}{\rm{{\cal I}}}\left\{ {j \ne i} \right\}
\end{array}
\end{align}

\begin{align}
\frac{{\partial x_j^{(L)}}}{{\partial w_{ij}^{(L)}}} = a_i^{(L - 1)}.
\end{align}

Therefore, $\frac{{\partial J(\omega ;\theta )}}{{\partial w_{ij}^{(L)}}} $ is derived by

\begin{align}
\begin{array}{l}
\frac{{\partial J(\omega ;\theta )}}{{\partial w_{ij}^{(L)}}} = ( - {\rm{{\cal I}}}\{ {S_k} = {S_j}\} (1 - a_j^{(L)}) + \sum\limits_{i = 0,i \ne j}^{{d^{(L)}}} {{\rm{{\cal I}}}\{ {S_k} = {S_i}\} } a_j^{(L)})a_i^{(L - 1)}\\
{\kern 1pt} {\kern 1pt} {\kern 1pt} {\kern 1pt} {\kern 1pt} {\kern 1pt} {\kern 1pt} {\kern 1pt} {\kern 1pt} {\kern 1pt}  {\kern 1pt} {\kern 1pt} {\kern 1pt}  {\kern 1pt} {\kern 1pt} {\kern 1pt} {\kern 1pt} {\kern 1pt} {\kern 1pt} {\kern 1pt} {\kern 1pt} {\kern 1pt} {\kern 1pt} {\kern 1pt} {\kern 1pt} {\kern 1pt} {\kern 1pt} {\kern 1pt}  = ({\rm{{\cal I}}}\{ {S_k} = {S_j}\} (a_j^{(L)} - 1) + (1 - {\rm{{\cal I}}}\{ {S_k} = {S_j}\} )a_j^{(L)})a_i^{(L - 1)}\\
{\kern 1pt} {\kern 1pt} {\kern 1pt} {\kern 1pt} {\kern 1pt} {\kern 1pt}  {\kern 1pt} {\kern 1pt} {\kern 1pt} {\kern 1pt} {\kern 1pt} {\kern 1pt} {\kern 1pt} {\kern 1pt} {\kern 1pt} {\kern 1pt}  {\kern 1pt} {\kern 1pt} {\kern 1pt} {\kern 1pt} {\kern 1pt} {\kern 1pt} {\kern 1pt} {\kern 1pt} {\kern 1pt} {\kern 1pt} {\kern 1pt} {\kern 1pt}  = (a_j^{(L)} - {\rm{{\cal I}}}\{ {S_k} = {S_j}\} )a_i^{(L - 1)}
\end{array}
\end{align}

Let $\delta _j^{(L)}=a_j^{(L)} - {\rm{{\cal I}}}\{ {S_k} = {S_j}\}$, the derivative can be simplified as
\begin{align}
\frac{{\partial J(\omega ;\theta )}}{{\partial w_{ij}^{(L)}}} = \delta _j^{(L)}a_i^{(L - 1)}
\label{equ19}
\end{align}

For the hidden layer $1 \le l \le L - 1$,
\begin{align}
\begin{array}{l}
\frac{{\partial J(\omega ;\theta )}}{{\partial w_{ij}^{(l)}}} = \sum\limits_{i = 0}^{{d^{(l)}}} {\frac{{\partial J(\omega ;\theta )}}{{\partial a_i^{(l)}}}\frac{{\partial a_i^{(l)}}}{{\partial x_j^{(l)}}}} \frac{{\partial x_j^{(l)}}}{{\partial w_{ij}^{(l)}}}\\
{\kern 1pt} {\kern 1pt} {\kern 1pt} {\kern 1pt} {\kern 1pt} {\kern 1pt} {\kern 1pt} {\kern 1pt} {\kern 1pt} {\kern 1pt} {\kern 1pt} {\kern 1pt} {\kern 1pt} {\kern 1pt} {\kern 1pt} {\kern 1pt} {\kern 1pt} {\kern 1pt} {\kern 1pt} {\kern 1pt} {\kern 1pt} {\kern 1pt} {\kern 1pt} {\kern 1pt} {\kern 1pt} {\kern 1pt} {\kern 1pt} {\kern 1pt}  = \sum\limits_{i = 0}^{{d^{(l + 1)}}} {\delta _i^{(l + 1)}w_{ij}^{(l + 1)}\tanh'(x_j^{(l)})a_i^{(l - 1)}}
\end{array}
\label{equ20}
\end{align}

where $\tanh'$ denotes the derivative of $\tanh$ function and the term $\delta _j^{(l)} = \frac{{\partial J(\omega ;\theta )}}{{\partial x_j^{(l)}}}$, which satisfies

\begin{align}
\delta _j^{(l)} = \sum\limits_{i = 0}^{{d^{(l + 1)}}} {\delta _i^{(l + 1)}w_{ij}^{(l + 1)}{{\tanh }^\prime }(x_j^{(l)})}.\label{equ21}
\end{align}

According to the equation (\ref{equ21}), all the delta values $\delta _j^{(l)}$ can be calculated recursively by the $\delta _i^{(l+1)}$ from the previous layers. Afterwards, the weights can be updated by using gradient descent with a learning rate \(\eta\),

\begin{align}
w_{ij}^{(l)} = \left\{ \begin{array}{l}
w_{ij}^{(l)} - \eta \sum\limits_{i = 0}^{{d^{(l + 1)}}} {\delta _i^{(l + 1)}w_{ij}^{(l + 1)}\tanh'(x_j^{(l)})a_i^{(l - 1)},{\kern 1pt} {\kern 1pt} {\kern 1pt} {\kern 1pt} {\kern 1pt} 1 \le l \le L - 1} \\
w_{ij}^{(l)} - \eta \delta _j^{(L)}a_i^{(L - 1)},{\kern 1pt} {\kern 1pt} {\kern 1pt} {\kern 1pt} {\kern 1pt} {\kern 1pt} l = L
\end{array} \right.
\label{equ22}
\end{align}

\section{Stacked BICM-ID Receiver with NN }

\begin{figure}[!t]
		\setlength{\abovecaptionskip}{0pt}
		\setlength{\belowcaptionskip}{0pt}
		\centering
		\includegraphics[ width=90mm]{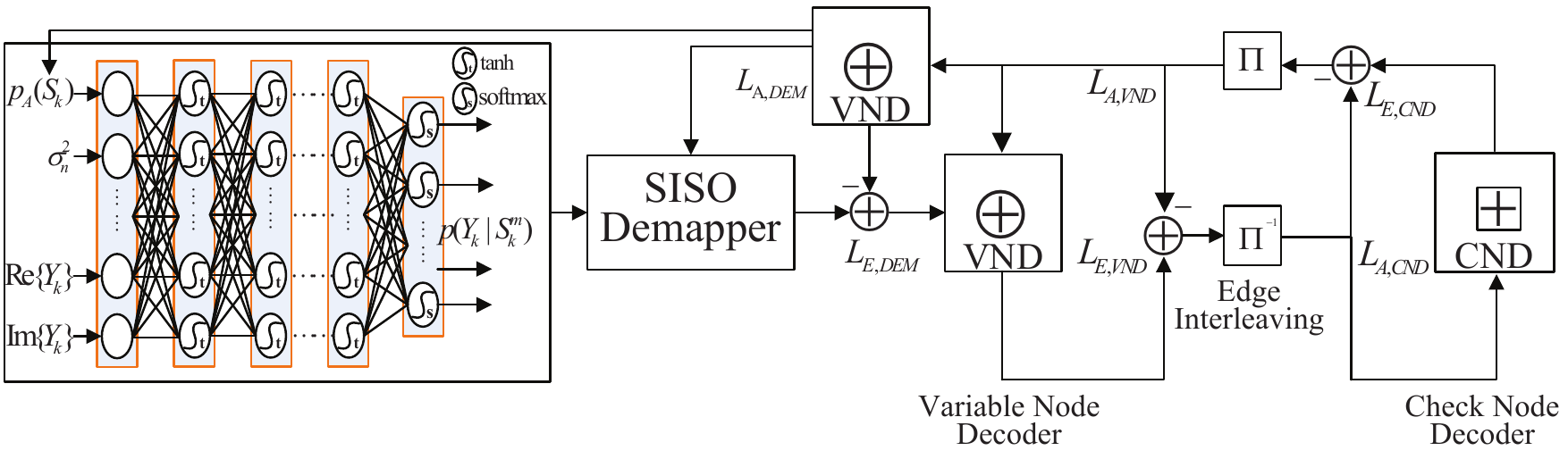}
		\caption{Stacked LDPC coded BICM-ID receiver with the NN.}
		\label{fig1b} \end{figure}

The LDPC coded BICM-ID receiver can be viewed as an iterative architecture with an inner MAP demodulator and an outer LDPC decoder. The LLR ${L_{{\rm{E,DEM}}}}(S_k^m)$ on the \(m\)-th bit is calculated and fed into the decoder to estimate the transmitted message bit  by equation
\begin{align}
\begin{array}{l}
{L_{{\rm{E,DEM}}}}(S_k^m) = \log \frac{{\sum\limits_{S_k^m \in \chi _1^m} {p({Y_k}|{S_k})\prod\nolimits_{m' = 1,m' \ne m}^M {p(S_k^{m'})} } }}{{\sum\limits_{S_k^m \in \chi _0^m} {p({Y_k}|{S_k})\prod\nolimits_{m' = 1,m' \ne m}^M {p(S_k^{m'})} } }}
\end{array}\label{equ23}
\end{align}
where the  $p(S_k^{m'})$ refers to the \(a\) \(priori\) probability of the set of symbols \(S_k\) taking the same \(m\)-th bit.

It has been claimed that the BICM-ID receiver can not achieve a further performance gain with the Gray labelling. To address this challenge, we propose a stacked BICM-ID design with the feed-forward NN, as shown in Fig. \ref{fig1b}. Taking the hybrid NN-aided BICM scheme at the first iteration, the stacked NN includes the  \(a\) \(priori\) probability $p(S_k)$ to the input layer additionally. The \(a\) \(priori\) probability $p(S_k)$ can be given by
\begin{align}
p({S_k}) = \prod\limits_{m = 1}^M {\frac{1}{2}(1 + {\rm tanh}(\frac{{L_{A,DEM}^{k,m}}}{2})S_k^m)}
\label{equ24}
\end{align}
where $L_{A,DEM}^{k,m}$ denotes the LLR \({L_{A,DEM}}\) corresponding to the \(m\)-th bit on the constellation \(S_k\) \cite{Xiaodong1999}. With the nonlinear clipping distortion, this stacked architecture can output the condition probability $p({Y_k}|S_k)$ at the \(i\)-th iteration based on the \(a\) \(priori\) probability $p(S_k)$ at the \((i-1)\)-th iteration, resulting in an iterative improvement on the calculation of the extrinsic LLR \({L_{E,DEM}}\). Since the NN changes with the \(a\) \(priori\) probability $p(S_k)$ at each iteration, we should customize the specific architecture in a stacked fashion from the NN at the last iteration. In the training procedure, we should search the appropriate hidden layer and training \(E_b/N_0\) \(\gamma_e \) separately at each iteration, whose the complexity will linearly increase with the iteration nummber.

\section{Simulation Results}
In this section, we present the numerical results of the NN-aided BICM receiver in the LDPC coded DCO-OFDM systems, where the parameters are shown in Table \ref{Tab1}. The channel bandwidth occupies \(N\) subcarriers with Hermitian symmetry, resulting in  \({N/2}-1\) information-carrying subcarriers. Given the Gray labelling, we select a \(2^M\)-ary quadrature amplitude modulation (QAM) constellations combined with a rate-\(R\) LDPC code as the coding and modulation scheme. Here, we consider the structured LDPC codes in the IEEE 802.11 protocol (WIFI-R1/2) \cite{IEEE802.11}, where the coded length is set to \(N_c\). The belief propagation decoding is used and the maximum number of iterations is set to 50.

We adopt the double hard clipping to fit the linear dynamic range of LED, whereas the nonlinear transfer characteristic  can be compensated by the pre-distortion. In the following, the optimal DC bias \(\mu \) that maximizes SNDR is chosen as the midpoint $\mu  = \frac{1}{2}({\Omega _b} + {\Omega _t})$ to balance the dynamic region, where the ${\Omega _b}$ equals to zero and ${\Omega _t}$ are selected according to the clipping level $\Psi$ \cite{Ying2015}. Here, the parameter $\Psi$ of the nonlinear distortion is evaluated by
\begin{align}
\Psi  = 10{\log _{10}}(\frac{{\Omega _t^2}}{{E(s_n^2)}}).
\end{align}
Specifically, the clipping level $\Psi$ of the 16-QAM and 64-QAM are respectively set to 9 dB and 11 dB. The bit error rate (BER) curves are plotted versus electrical \(E_b/N_0\) $\gamma_e$, denoted as
\begin{align}
\gamma_e = 10\lg \left(\frac{{{P_e}(\delta _b,\delta _t ,{\sigma _s})}}{{2\varepsilon MR \sigma _n^2}}\right)
\end{align}
where the bandwidth utilization factor \(\varepsilon\) is denoted by $\varepsilon  = \frac{1}{2} - \frac{1}{N}$ in DCO-OFDM.

\begin{table}[!t]
\renewcommand{\arraystretch}{1}
\caption{The parameters of the NN-aided BICM design for LDPC coded DCO-OFDM }
\label{Tab1}
\centering
\begin{tabular}{|c|c|c|c|}
\hline
$\bf{System \ Model}$ & $\bf{Values}$  \\
\hline
Code Length $N_c$ & 1296/1944/2304  \\
\hline
Code Rate & 1/2  \\
\hline
Modulation & QPSK 16-QAM 64-QAM 256-QAM  \\
\hline
Labelling & Gray  \\
\hline
IFFT/FFT Size & 64/256/1024  \\
\hline
CL & 9dB(16-QAM) 10dB(64-QAM) \\
\hline
$\bf{Neural \ Network}$ &   \\
\hline
Loss Function & cross-entropy \\
\hline
Gradient Descent optimization & Scaled Conjugate Gradient  \\
\hline
Training Dataset & 50$N_c$  \\
\hline
Training $E_b/N_0$ & \tabincell{c}{FFT-64: 13dB(16-QAM) 10dB(64-QAM) \\ FFT-1024: 10dB(16-QAM) 9dB(64-QAM)} \\
\hline
Hidden Layers & \tabincell{c}{NN1: $[$32 16 8$]$, NN2: $[$128 64 32$]$ \\ NN3: $[$128 64 32 16$]$, NN4: $[$512 128 64 32$]$  }  \\
\hline

\end{tabular}
\end{table}

We present the specific training parameters of the NN design, such as the size of training dataset, the training \(E_b/N_0\) $\gamma_{t}$ and the hidden layers etc., as shown in Table \ref{Tab1}. There are 4 network architectures with different hidden layers, which are denoted as NN1-NN4. It indicates that the NN-aided BICM works well after training a small sample dataset with \(50N_c\) bits. To investigate the impact of the NN on the performance of BICM receiver, we pick up the robust feed-forward architecture and the appropriate training \(E_b/N_0\) values $\gamma_{t}$ for the corresponding coded modulation schemes from the candidate options in the subsection of Neuaral Network. In the following simulations, we compare our proposed NN-BICM receiver with the MAP-BICM, GMM-BICM and MLSD-BICM counterparts. Different from the GMM-BICM using EM algorithm, NN-BICM trains only once on the appropriate $\gamma_{t}$ rather than training the related parameters corresponding to each \(E_b/N_0\) value.

\begin{figure}[!t]
		\setlength{\abovecaptionskip}{0pt}
		\setlength{\belowcaptionskip}{0pt}
		\centering
		\includegraphics[ width=81mm]{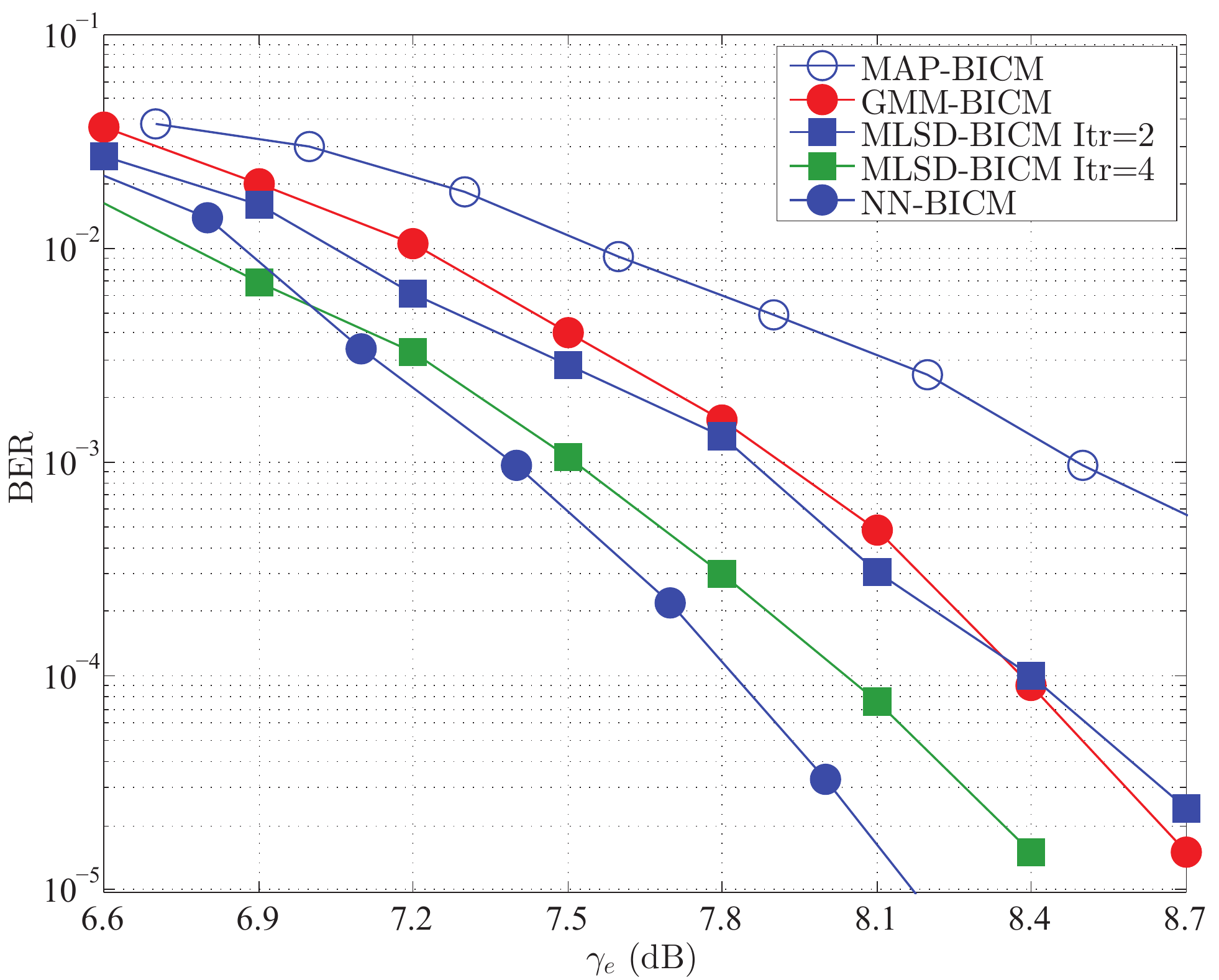}
		\caption{BER performance of the NN-aided BICM in the LDPC coded DCO-OFDM systems with 16-QAM (FFT-64).}
		\label{fig4} \end{figure}

\begin{figure}[!t]
		\setlength{\abovecaptionskip}{0pt}
		\setlength{\belowcaptionskip}{0pt}
		\centering
		\includegraphics[ width=85mm]{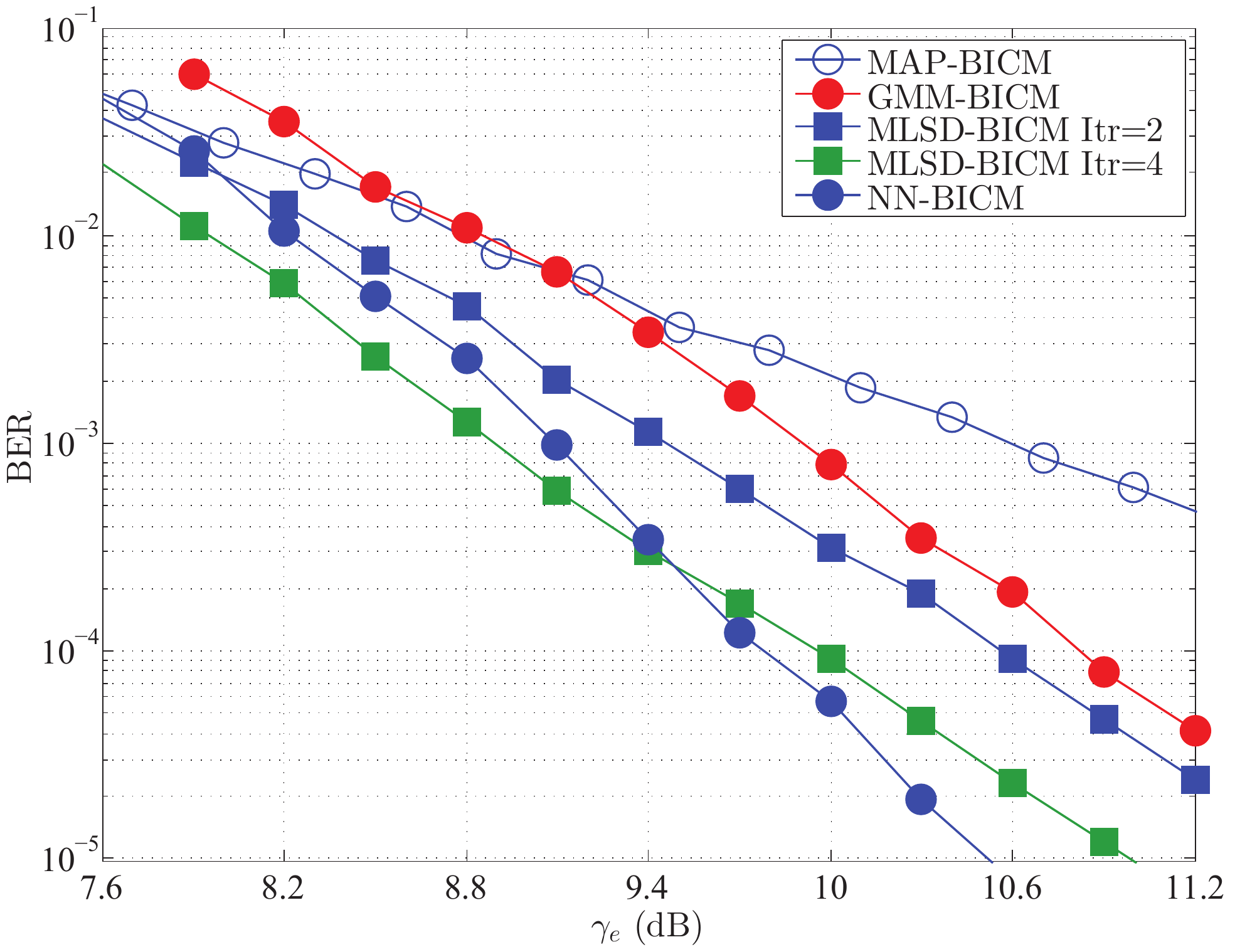}
		\caption{BER performance of the NN-aided BICM in the LDPC coded DCO-OFDM systems with 64-QAM (FFT-64).}
		\label{fig5} \end{figure}

\begin{figure}[!t]
		\setlength{\abovecaptionskip}{0pt}
		\setlength{\belowcaptionskip}{0pt}
		\centering
		\includegraphics[ width=85mm]{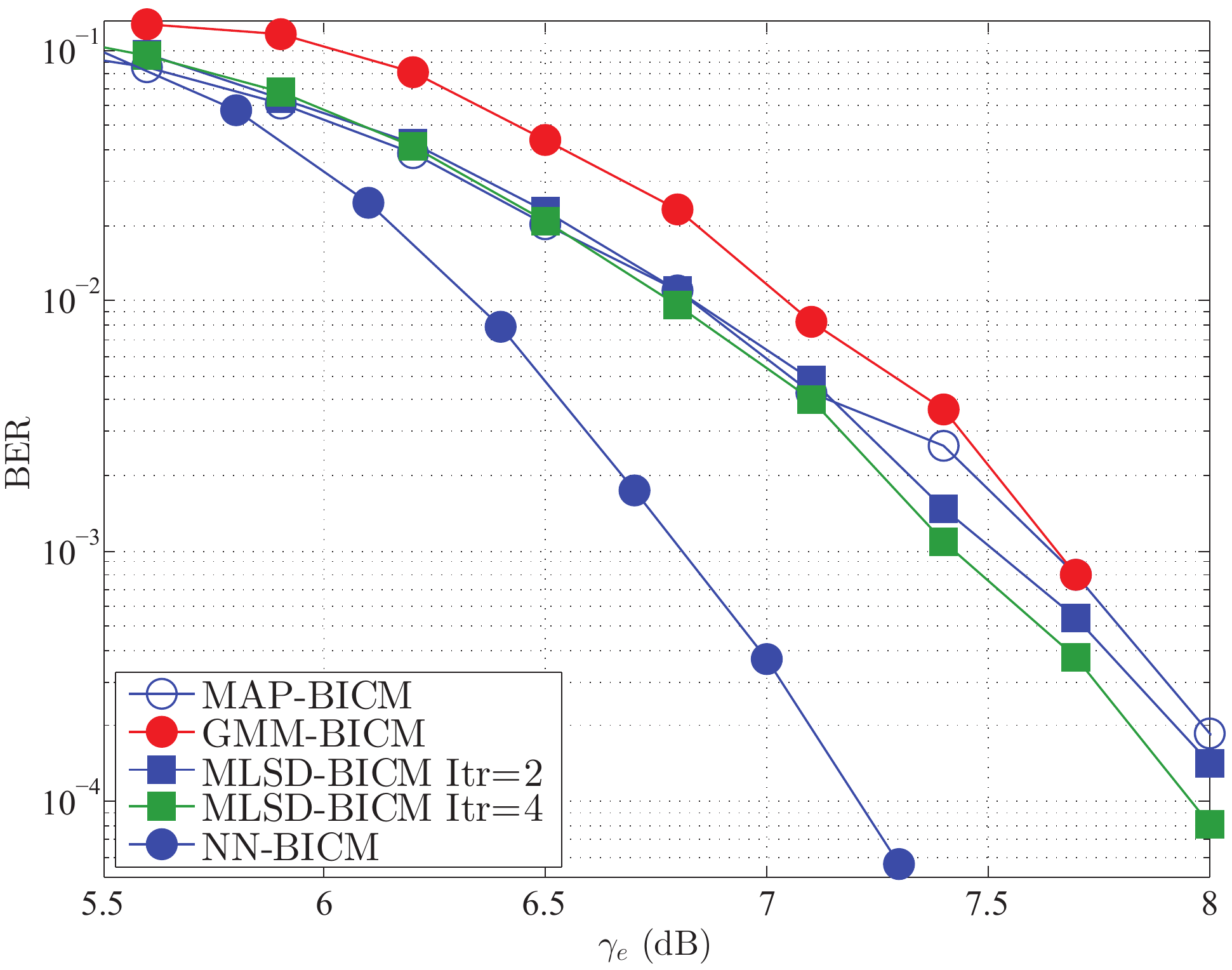}
		\caption{BER performance of the NN-aided BICM in the LDPC coded DCO-OFDM systems with 16-QAM (FFT-1024).}
		\label{fig6} \end{figure}

Fig. \ref{fig4} illustrates the BER comparisons of the NN-aided BICM receiver and other receiver schemes. First, we consider an LDPC coded DCO-OFDM system with 64 subcarriers and 16-QAM. In the NN-BICM design and the GMM-BICM design, we choose the NN with the hidden layers [32 16 8] and training Eb/N0 $\gamma_{t}=13$ dB. The curve of the MAP-BICM is given as reference. However, there exists a noticeable gain on the performance of the MLSD-BICM with two and four iterations, which indicates that MLSD-BICM can achieve an additional improvement with the increasing iterations when combating the nonlinear distortion. Eventhough, it can be observed that the NN-BICM design clearly outperforms the MLSD-BICM with 4 iterations, MAP-BICM and GMM-BICM by about 0.2 dB, 0.5 dB and 1.5 dB respectively at a BER of 1e-4, which demonstrates the superiority of the NN architecture.

In Fig. \ref{fig5}, the BER curves of the the NN-aided BICM receiver and other designs for the LDPC coded DCO-OFDM system with 64-QAM and the 64 subcarriers are presented. The NN with the hidden layers [32 16 8] and training Eb/N0 $\gamma_{t}=15$ dB is considered. BER results show that all the curves are becoming slow down when the nonlinear distortion imposes a strong impact on the higher order modulation. Specifically, NN-BICM provides a remarkable performance gain than GMM-BICM and MAP-BICM by about 1.2 and 1.8 dB respectively at a BER of 1e-3. Moreover, we can see that the MLSD-BICM with 4 iterations exhibits better performance than the NN-BICM at low Eb/N0 region from 7.5 to 9.5 dB, while it suffers at high Eb/N0 region that larger than 9.5 dB. The NN-BICM outperforms MLSD-BICM by about 0.5 dB at 1e-5 as the Eb/N0 $\gamma_{e}$ increases, which can verify the benefits of the NN design.

\begin{figure}[!t]
		\setlength{\abovecaptionskip}{0pt}
		\setlength{\belowcaptionskip}{0pt}
		\centering
		\includegraphics[ width=81mm]{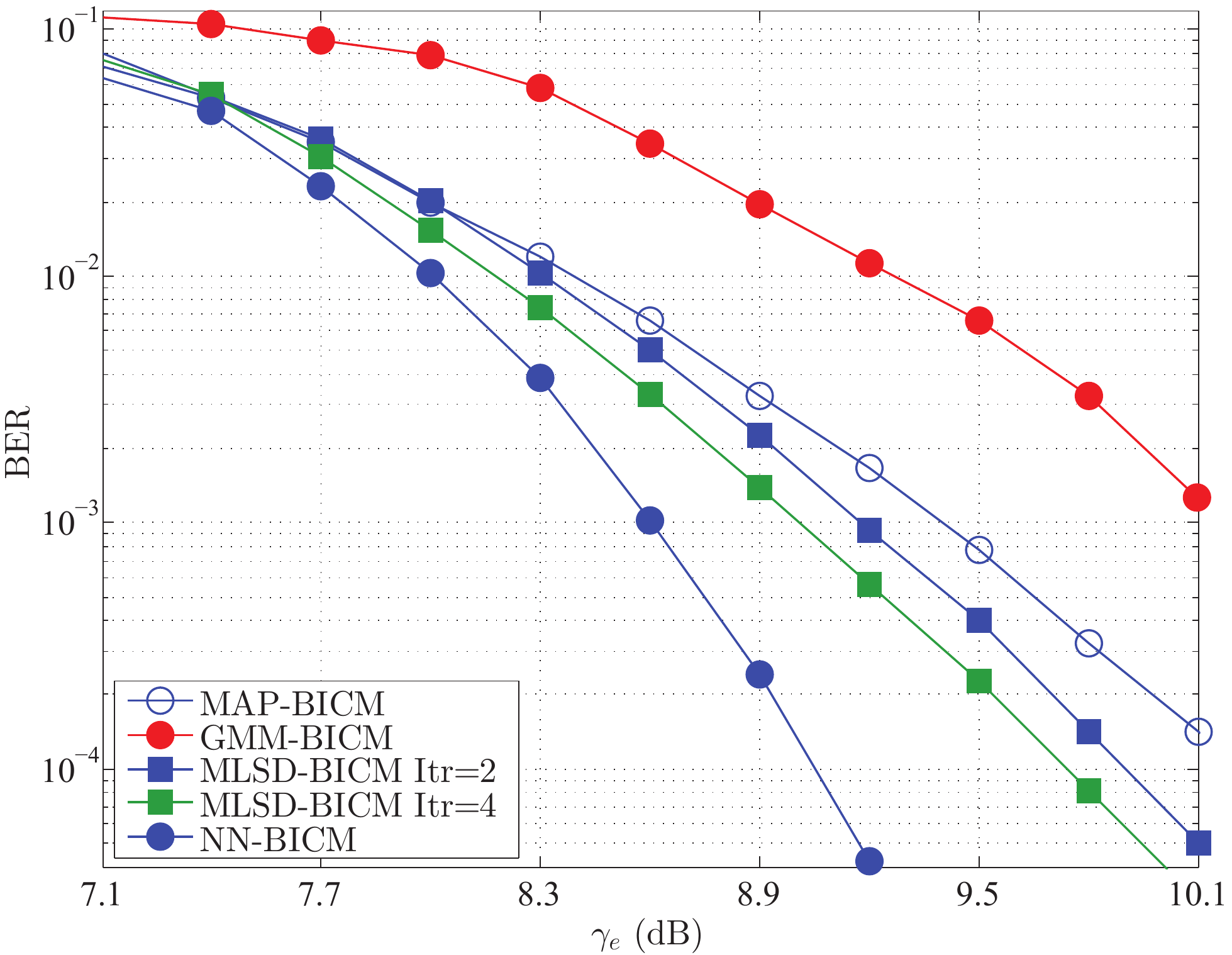}
		\caption{BER performance of the NN-aided BICM in the LDPC coded DCO-OFDM systems with 64-QAM (FFT-1024).}
		\label{fig7} \end{figure}

Fig. \ref{fig6} shows the BER performance of the 16-QAM modulation between the NN-aided BICM receiver and other designs with 1024 subcarriers. The NN with the hidden layers [32 16 8] and training Eb/N0 $\gamma_{t}=13$ dB is adopted. It is obvious that NN-BICM achieves a noticeable performance gain by about 0.8 to 1 dB over the other counterparts at a BER of 1e-4. We observe that neither GMM-BICM nor the MLSD-BICM can improve the performance gain in this situation that the LDPC coded DCO-OFDM system with 1024 subcarriers suffers from the the clipping level of 9 dB. GMM-BICM fails modeling the mix-Gauss distribution, since the conditional probability $p({Y_k}|S_k;\theta)$ approach to be Gaussian distributed when the subcarriers \(N \to \infty \). While the MLSD-BICM struggles to improve the gain when the MLSD has the difficulty in searching the optimal solution over the space of \(M^N\) possible candidate symbol sequence with an exponential growth of the increasing \(N\). In Fig. \ref{fig7}, the superiority for the NN-aided BICM is more evident with the higher modulation order. We can see that the GMM-BICM exhibits a remarkable deterioration about 0.6 dB performance gap, in comparison with the MAP-BICM scheme.

\begin{figure}[!t]
		\setlength{\abovecaptionskip}{0pt}
		\setlength{\belowcaptionskip}{0pt}
		\centering
		\includegraphics[ width=81mm]{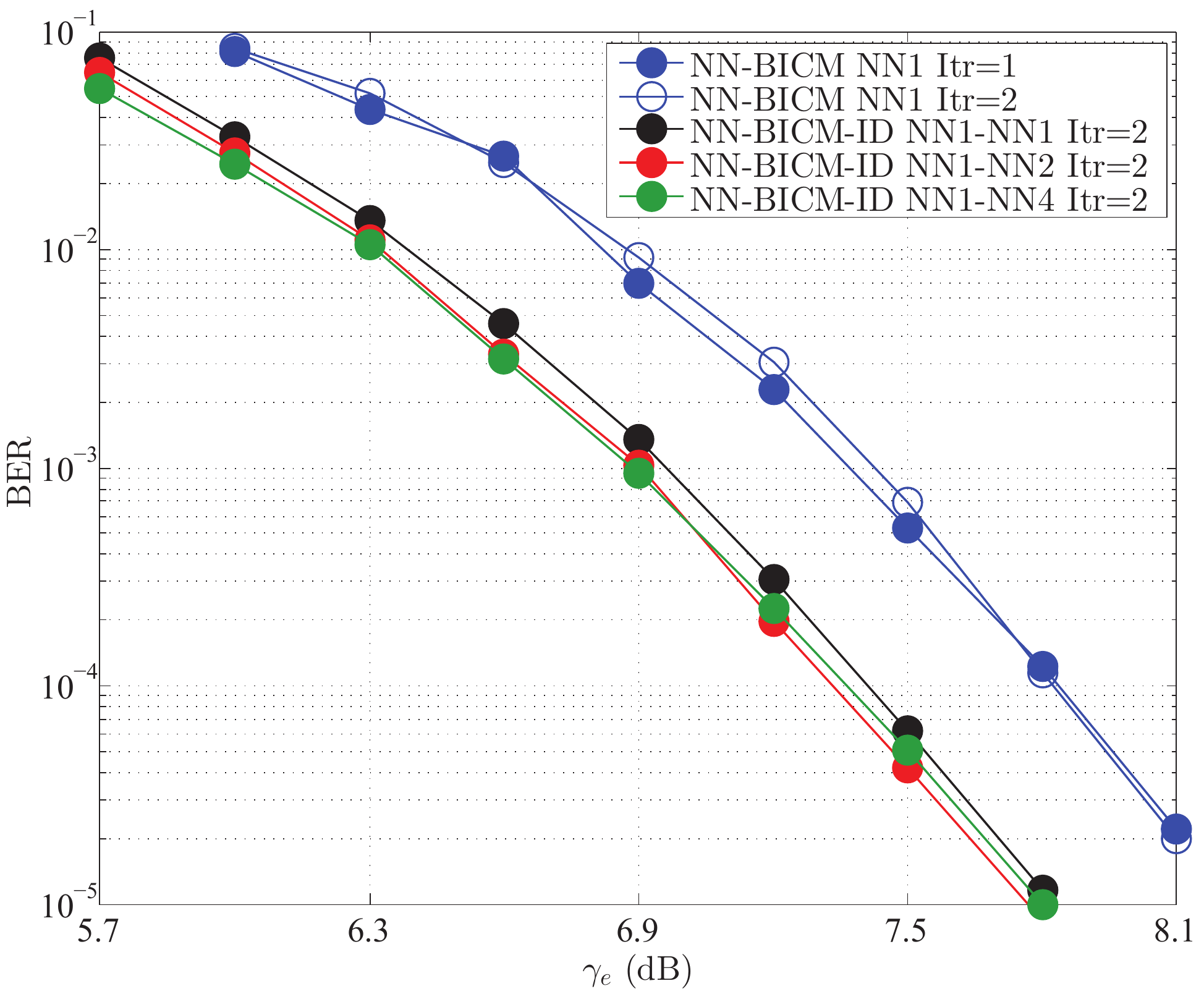}
		\caption{BER performance of the NN-aided BICM-ID in the LDPC coded DCO-OFDM systems with 16-QAM (FFT-64).}
		\label{fig8} \end{figure}

Fig. \ref{fig8} depicts the BER comparisons between the NN-BICM and stacked NN-BICM-ID schemes when adopting 64 subcarriers and 16-QAM in the LDPC coded DCO-OFDM system with the clipping level 9dB. In NN-BICM, the hidden layer [32 16 8] and training Eb/N0 \(\gamma_t\)=13 dB are considered. The NN-BICM scheme gets the channel conditional probability through the feed-forward network and calculates modified LLRs \({L_{E,DEM}}\) by the soft demapper. Traditionally, NN-BICM can feed the \(a\) \(priori\) knowledge in the decoder back to the demapper iteratively to develop a candidate iterative method. However, as depicted in Fig. \ref{fig8}, we find that the Gray mapping can not achieve a further performance gain with the increasing iterations by using this method, whereas the performance of NN-BICM with 2 iterations is similar to the first iteration. On the other hand, the stacked NN-BICM-ID uses the same parameters as NN-BICM at the first iteration, and employs the different hidden layers and the training Eb/N0 \(\gamma_t\)=10 dB at the second iteration. It shows that NN-BICM-ID, which employs the architecture NN1 and NN2 at the each iteration time respectively, outperforms NN-BICM by about 0.6 dB at a BER of 1e-5, which exhibits a significant performance gain with the same iteration time. Besides, NN-BICM-ID with the architecture of NN1 and NN4 provides a similar performance as that of NN1 and NN2 at the cost of the extra hidden neurons.

\section{Conclusion}

In this paper, we introduce a novel NN-BICM and NN-BICM-ID receiver in the LDPC coded DCO-OFDM system respectively. These feed-forward networks are simplified by establishing the input layer with a single symbol and the corresponding variance of the Gaussian noise, in addition with a feed back of the \(a\) \(priori\) probability from the LDPC decoder for the NN-BICM-ID design alternatively. Both revised BICM/BICM-ID receivers can improve the LLR values by adopting the loss function of cross-entropy and the $softmax$ activation function. The numerical results show that NN-BICM and NN-BICM-ID can provide a better BER performance than other counterparts.

\end{document}